%% file: main.tex
\documentclass[runningheads]{llncs}

\usepackage[review,year=2026]{eccv}

\usepackage{eccvabbrv}
\usepackage{graphicx}
\usepackage{booktabs}
\usepackage[accsupp]{axessibility}

\usepackage[pagebackref,breaklinks,colorlinks,citecolor=eccvblue]{hyperref}
\usepackage{orcidlink}

\usepackage{subfiles}
\usepackage{multirow}
\usepackage{tikz}
\usepackage{enumitem}
\usepackage{amsmath}
\newcommand{\circlednum}[1]{%
  \tikz[baseline=(char.base)]{
    \node[shape=circle,fill=blue!20,inner sep=1pt] (char) {\textbf{#1}};%
  }%
}

\begin{document}
\nolinenumbers

\title{TraversRL: Traversable Pedestrian Pathway Generation With Reinforcement Learning} 

\titlerunning{Traversable Pedestrian Pathway Generation}


\author{Bin Han\inst{1}\orcidlink{0000-0002-5280-9456} \and
Robert Wolfe\inst{2}\orcidlink{0000-0001-7133-695X} \and
Bill Howe\inst{1}\orcidlink{0000-0001-8588-8472}}

\authorrunning{B.~Han et al.}

\institute{University of Washington, Seattle, WA, USA \and
Rutgers University, New Brunswick, NJ, USA \\
\email{bh193@uw.edu, robert.wolfe@rutgers.edu, billhowe@uw.edu}}

\maketitle

\begin{abstract}
Automatically generating pedestrian pathways from aerial images requires producing a connected network suitable for routing, not just detecting where sidewalks appear. Sidewalks and crossings, in contrast to roads, may be partially occluded, implicitly defined, and exhibit complex connectivity patterns. Existing segmentation-based approaches focus on labeling pixels to infer segments, but often produce disconnected or fragmentary graphs that are unreliable for navigation. We introduce \textbf{TraversRL}, a vision-conditioned model that iteratively grows a pathway network from an aerial image, simulating a traveler navigating the built environment. TraversRL uses an action space of short and long direction–distance segments designed to adapt to complex patterns and span occlusions, and uses a combination of graph-level and step-wise rewards to balance overall connectivity with precise edge placement. Across three visual backbones and three intersection datasets, TraversRL substantially improves buffered IoU with the ground-truth graph relative to a state-of-the-art segmentation baseline, and more than doubles metrics of connectivity. Moreover, combining global and local rewards produces cleaner graphs with fewer spurious branches while further improving overall performance. These results demonstrate that modeling pathway extraction as a sequential decision process from the perspective of a traveler, while optimizing for final graph quality with reinforcement learning, produces significantly more reliable pedestrian networks.

\end{abstract}

\section{Introduction}\label{sec:intro}
\subfile{sections/introduction}

\section{Related Work}\label{sec:related}
\subfile{sections/related_work}

\section{Method}\label{sec:method}
\subfile{sections/method}

\section{Experiment}\label{sec:experiment}
\subfile{sections/experiment}

\section{Results}\label{sec:results}
\subfile{sections/results}

\section{Conclusion}\label{sec:conclusion}
\subfile{sections/conclusion}

\section*{Acknowledgements}\label{sec:acknowledgement}

We thank the Taskar Center for Accessible Technology for allowing us to derive the WashingtonInter dataset from the OS-CONNECT project.

\newpage
\bibliographystyle{splncs04}
\bibliography{main}

\end{document}

%% file: sections/introduction.tex
Accurate maps of pedestrian pathways are foundational for accessible navigation, mobility analysis, and urban planning. Yet these data are difficult to collect, and often incomplete or inaccurate, creating failure cases that disproportionately affect pedestrians and accessibility-critical users~\cite{zhang2023ape}. Inferring connected, traversible networks of sidewalks and crossings is a challenging vision problem: paths vary in width, shape, and material; they frequently occluded by trees in overhead imagery; visual cues can be subtle or absent; and small geometric errors can induce large topological failures that degrade end-to-end routability (Fig.~\ref{fig:teaser}). 

Existing mapping pipelines rely heavily on segmentation: predicting pixel-level raster labels followed by post-processing. However, segmentation-based methods tend to produce fragmented graphs that are unreliable for navigation, as paths may not be visible from overhead, and travelers may cross the street at unlabeled crosswalks or cut through parking lots. An alternative approach is iterative graph construction, where a model repeatedly extends a partial network conditioned on local image information and the generated context, simulating a traveler making step-by-step decisions to navigate the built environment. 

RL fine-tuning has played an increasingly important role for multi-step problem solving, particularly in settings where intermediate supervision is ambiguous but final outcomes are more easily scored~\cite{openai2024o1systemcard,guo2025deepseek,team2025kimi,gao2024designing,wen2025reinforcement,wang2025reinforcement}. This paradigm closely matches pathway generation: connectivity can be guaranteed step by step, and overall quality is measured on the final graph since small errors can accumulate.

We introduce \textbf{TraversRL}, a vision-conditioned generator that iteratively constructs pedestrian graphs from aerial imagery. It takes an intersection-level image crop and a partial graph to predict the next network extension, using a discretized direction--distance action space that includes both short and longer distance segments. We train TraversRL in two stages: \textbf{\circlednum{1}} \textit{Supervised pretraining.} The model learns to imitate ground-truth graph expansions step by step via cross-entropy loss, augmented with localization noise to improve robustness against drift during iterative rollouts. We denote this pretrained model as TraversRL-Pre. \textbf{\circlednum{2}} \textit{RL fine-tuning.} Initializing from TraversRL-Pre as a reference policy, we optimize outcome-level graph rewards under a KL-regularized objective, implemented via group-relative policy optimization (GRPO)~\cite{shao2024deepseekmath,guo2025deepseek}.

We investigate two RL reward designs. \textbf{\circlednum{1}} \textit{Global reward} uses a terminal, graph-level objective based on buffered geometry intersection-over-union (IoU), which aligns with final network quality. We refer to this model as TraversRL-Global. \textbf{\circlednum{2}} \textit{Local stepwise reward} augments the terminal objective with per-edge buffered overlap efficiency to encourage locally precise geometry.  We refer to this model as  TraversRL-Local. Overall, our main contributions are:
\begin{figure}[!t]
    \centering
    \includegraphics[width=\columnwidth]{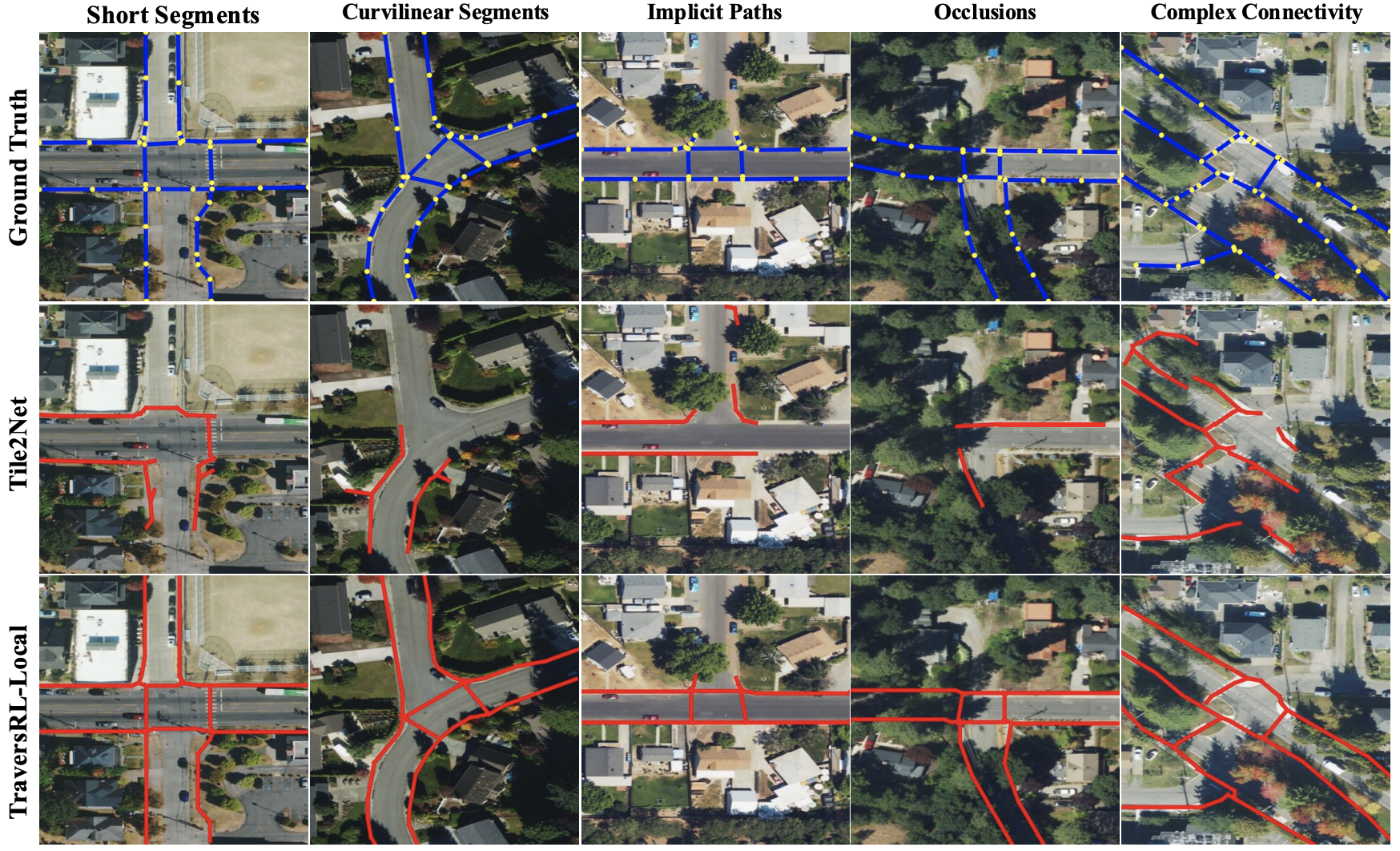}
    \caption{\textbf{Qualitative examples of pedestrian pathway generation.} Representative aerial intersection scenes illustrating common challenges for pedestrian mapping: \{short segments, curvilinear segments, implicit paths (no crossing marks), occlusions from trees, and complex connectivity\}. \textbf{Top}: ground-truth graph (blue; yellow dots mark vertices). \textbf{Middle}: predictions from the state-of-the-art segmentation-based method, Tile2Net (red). \textbf{Bottom}: preductions from our TraversRL-Local model (red). These examples highlight the topology-sensitive nature of pedestrian networks, and demonstrate TraversRL’s robust generation across diverse real-world challenges.}
    \label{fig:teaser}
\end{figure}

\begin{itemize}
    \item We introduce TraversRL, a vision-conditioned generator that iteratively extracts complex pathway graphs from aerial imagery, using a discretized direction–distance action space tailored to more complicated pedestrian segments.
    
    \item We demonstrate that iterative generation with both supervised pre-training and RL fine-tuning is a natural fit for pathway tracing and consistently improves graph quality. Averaged across three backbones and three datasets, TraversRL-Global outperforms a state-of-the-art segmentation-based method in IoU by 43.6\% and a sidewalk connectivity metric (TravSim) by 205\%. 
    
    \item We propose a local stepwise reward that measures geometric quality for advantage shaping during RL fine-tuning. On average, TraversRL-Local further improves IoU by 2.6\% and TravSim by 1.6\% relative to TraversRL-Global.
    
    \item We show that RL fine-tuning generalizes zero-shot to unseen cities without additional training. RL fine-tuning also yields cleaner and simpler graphs by reducing redundant edges while maintaining connectivity for navigation.
\end{itemize}

%% file: sections/related_work.tex
\paragraph{Street and Road Network Generation.}
Road and street network extraction from aerial imagery has been studied extensively. Existing methods broadly fall into two paradigms. \textit{Segmentation-first} pipelines predict dense road masks and then recover graph structure through vectorization and post-processing~\cite{deng2025gld,hetang2024segment,xu2023road,he2020sat2graph,sun2019leveraging,zhou2021funet,lu2021gamsnet,mi2021hdmapgen}. In contrast, \textit{iterative graph construction} grows a graph step-by-step by conditioning on local image evidence and the already-generated context. RoadTracer~\cite{bastani2018roadtracer} introduced this crop-and-canvas tracing paradigm. VecRoad~\cite{tan2020vecroad} improved local exploration at complex junctions, while RNGDet and RNGDet++~\cite{xu2022rngdet,xu2023rngdet++} further developed transformer-based iterative graph detection with topology-oriented supervision. NETracer~\cite{liu2025netracer} proposed a topology-aware tracing framework by explicitly modeling node--edge connectivity.

While these approaches establish strong foundational techniques for extracting road graphs, they are not directly transferable to the pedestrian environment. Pedestrian pathways exhibit different challenges: they are typically narrower, contain many short and curvilinear segments, are more susceptible to occlusions (e.g., tree canopies), and require the explicit modeling of crossings and intersection connectivity to for accessibility and navigation use cases~\cite{hosseini2023mapping,zhang2024pathwaybench}. These inherent differences make pedestrian mapping more sensitive to rollout drift and localized geometric mispredictions during iterative generation.

\paragraph{Pedestrian Pathways From Aerial Imagery.}
Pedestrian pathway extraction from overhead imagery remains relatively understudied. Karimi et al.~\cite{karimi2013pedestrian} explored pedestrian map generation at small scale and highlighted the importance of input data quality. Tile2Net~\cite{hosseini2023mapping} detects sidewalks, crosswalks, and footpaths from aerial imagery and converts them into pedestrian networks. Prophet~\cite{zhang2023ape} combines road-network priors with aerial-image segmentation and rasterized street maps to infer pathway graphs (we do not compare with Prophet as neither the code nor trained model was available at the time of this writing). PathwayBench~\cite{zhang2024pathwaybench} argues that pixel-level metrics can understate mobility-critical errors and instead emphasizes routability and intersection-scale connectivity. Other work focuses on specific pedestrian assets rather than full connected networks, specifically zebra-crossing detection~\cite{ahmetovic2015zebra}, crosswalk detection from satellite imagery~\cite{verma2024crosswalk,bhuyan2025crosswalknet}, and curb-ramp detection from street-view imagery~\cite{hara2014tohme,o2025rampnet}. In contrast, our goal is direct generation of connected pedestrian pathway graphs.

\paragraph{Reinforcement Learning (RL) and GRPO.}
Reinforcement learning is increasingly used to improve long-horizon decision-making. Reasoning LLMs trained with RL show strong gains with verifiable rewards (RLVR), where policies are optimized using outcome-level signals because intermediate supervision is ambiguous~\cite{openai2024o1systemcard,guo2025deepseek,team2025kimi,gao2024designing,wen2025reinforcement,wang2025reinforcement}. More broadly, RL is a natural choice for optimizing non-differentiable trajectory-level objectives when errors can compound over many decisions. Among policy-gradient methods, PPO~\cite{schulman2017proximal} is widely used for its stability. GRPO~\cite{shao2024deepseekmath} is a PPO-style variant that replaces an explicit critic with group-relative baselines computed from multiple outputs, and typically uses KL regularization to a reference policy for stable updates. Our work uses GRPO.

These developments suggest RL is a natural fit for improving sequential generators when the desired objective is best expressed at the trajectory level. In pedestrian pathway generation, local geometric choices (e.g., sidewalk continuations and crossings) interact across many steps and must ultimately produce a globally coherent, routable network, while evaluation can be defined using geometry and connectivity aware graph metrics. We therefore adopt RL for fine-tuning and use GRPO as a practical, stable instantiation in our experiments.

%% file: sections/method.tex
We propose \textbf{TraversRL}, a vision-conditioned generator that constructs pedestrian pathway graphs step-by-step from aerial imagery. TraversRL is trained with two stages --- supervised pretraining (Sec.~\ref{sec:pretraining}), followed by RL fine-tuning using either a global graph-level reward (Sec.~\ref{sec:grpo_global}) or with an additional local stepwise reward (Sec.~\ref{sec:grpo_local}). The supervised pretraining phase yields a reference policy, denoted as TraversRL-Pre. During the subsequent RL fine-tuning, we initialize the trainable policy $\pi_\theta$ from TraversRL-Pre while maintaining a frozen copy $\pi_{\theta_0}$ to serve as the reference policy for KL divergence.

\subsection{Supervised Pretraining of TraversRL (TraversRL-Pre)}\label{sec:pretraining}

\textbf{Graph formulation.} Given a satellite image and vector pathway annotations, we convert the annotations into an undirected graph $\mathcal{G}=(V,E)$. Each node $v\in V$ is a 2D point in normalized coordinates $v=(x,y)\in[0,1]^2$, and each edge $e=\{u,v\}\in E$ is a LineString segment. Our goal is to learn a policy that incrementally reconstructs $E$ by predicting local edge continuations.

\textbf{State: aerial image crop + generation canvas.}  At each step, we expand from a current node $u$ using a local observation centered at $u$: (1) an RGB crop $I_u$ and (2) a single-channel generation canvas $C_u$ that rasterizes the partial predicted graph (all previously generated edges). Both are extracted with a fixed normalized window ($0.5\times0.5$ of the full extent) and resized to a fixed resolution ($256\times256$; $224\times224$ for ViT). We concatenate them into a 4-channel input $X_u=\mathrm{concat}(I_u,C_u)\in\mathbb{R}^{4\times H\times W}$, so that each decision is explicitly conditioned on local appearance and the previously generated edges.

\textbf{Action space: discretized direction and distance.} TraversRL predicts a discrete movement action from $u$ or \textsc{Stop}. For a neighbor $v$ of $u$, the displacement $\Delta=v-u$ is mapped to a class by discretizing angle and distance: $N_\theta=36$ angle bins (10$^\circ$) and $N_r=10$ distance bins over $[0,0.2]$ (normalized to the overall image), yielding $|\mathcal{A}|=N_\theta N_r + 1 = 361$ actions, where \textsc{Stop} indicates that no ungenerated edges remain at the current node. We denote $a\in\mathcal{A}$ as an action.

\textbf{Policy network.} TraversRL uses an ImageNet-pretrained backbone $f_\phi(\cdot)$ adapted to accept 4-channel inputs. We replace the first convolution with a 4-channel variant by copying the pretrained RGB weights and initializing the additional (canvas) channel to zero. Given $X_u$, the backbone produces a feature vector $h\in\mathbb{R}^d$, which is mapped to action probabilities by an MLP head:
\begin{equation}
\pi_\theta(a\mid X_u)=
\mathrm{softmax}\!\left(W_2\,\sigma\!\bigl(W_1\,\mathrm{Dropout}(h)\bigr)\right),
\end{equation}
where $\sigma(\cdot)$ is ReLU and $\pi_\theta(\cdot\mid X_u)\in\mathbb{R}^{|\mathcal{A}|}$.

\begin{figure}[!t]
    \centering
    \includegraphics[width=\columnwidth]{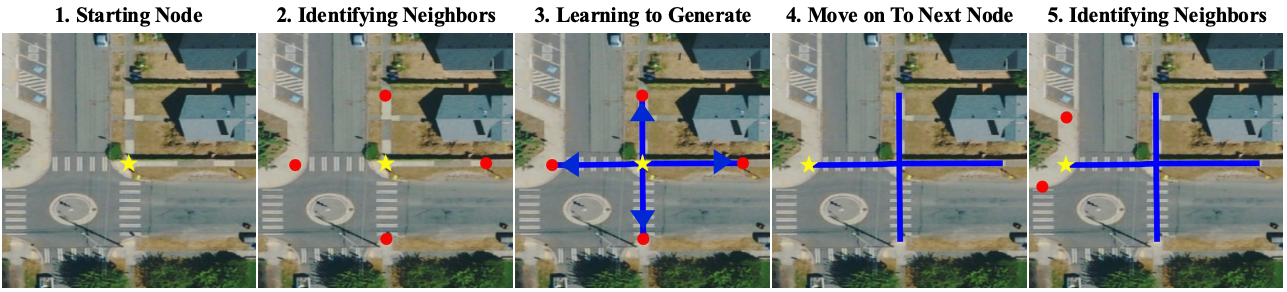}
    \caption{\textbf{Supervised node expansion in TraversRL pretraining.} Given the current node (yellow star), the set of ground-truth neighbors (red) are determined. During supervised pretraining, the policy is trained to sequentially emit one direction--distance action per ground-truth edge (blue arrows), adding the corresponding edges to the canvas. After all edges at the current node have been generated, the policy emits \textsc{Stop} and the algorithm moves to the next node for expansion (steps 1--5).}
    \label{fig:iterative}
\end{figure}

\textbf{Supervised training.} For each training graph, we supervise a sequence of actions at each node: the policy emits one action per ground-truth edge, then emits \textsc{Stop} and moves to the next node (Fig.~\ref{fig:iterative}). 
To improve robustness to rollout drift and mitigate error accumulation, we inject localization noise by perturbing the current node location while following the ground-truth edge:
\begin{equation}
\tilde{u}=u+\epsilon,\qquad \epsilon\sim \mathcal{U}\!\left([-\eta,\eta]^2\right).
\end{equation}
We extract crops centered at $\tilde{u}$ and compute the target class using the perturbed displacement $\Delta=v-\tilde{u}$
We minimize cross-entropy over the supervised action sequence.  The supervised training approach is adapted from a similar technique for road extraction~\cite{bastani2018roadtracer}, but includes variable distances and a different architecture to adapt to topologically complex, occluded, high-density sidewalk paths.

\subsection{Inference and Rollout Hyperparameters}\label{sec:inference}

At inference time, we roll out $\pi_\theta$ to construct a graph in continuous space. We initialize a start node (for training and evaluation, a random node in the ground-truth graph; in practice, a point selected by a user or based on segmentation pre-processing) and maintain a stack of active nodes. For each popped node $u$, we repeatedly call the policy until it emits \textsc{Stop} or reaches a per-node expansion budget. Each non-\textsc{Stop} action is decoded into a displacement using the corresponding angle--distance bin midpoints, creating a new node at $u+\Delta$ (clipped to $[0,1]^2$). We add the edge, render it onto the global canvas, and push the new node onto the stack. To avoid trivial cycling, we disallow executing the same discrete action twice from the same node within one expansion.

Because backbones can differ in termination behaviors, we tune a small set of rollout hyperparameters on the validation set. We configure the following:
\begin{enumerate}
    \item \emph{Top-$k$ backoff} ($k$): enumerate the $k$ highest-logit actions and execute the first geometrically valid action. $k=1$ equates to standard greedy decoding.
    \item \emph{Restarts per graph} ($R$): run the full rollout $R$ times from independently sampled start nodes and retain the best result under an internal confidence score (defined as the sum of top-1 minus top-2 logit margins across all steps).
    \item \emph{\textsc{Stop} threshold} ($\tau_{\text{stop}}$): accept \textsc{Stop} only when $\pi_\theta(\textsc{Stop}\mid s_t)\ge\tau_{\text{stop}}$.
    \item \emph{Maximum per-node expansions} ($M$): cap on the total number of actions permitted when expanding any single node.
\end{enumerate}
We select the configuration that maximizes buffered IoU on the validation set and reuse it for the subsequent RL training and inference.

\subsection{RL Fine-Tuning with Global Graph Reward (TraversRL-Global)}\label{sec:grpo_global}

\paragraph{Motivation.} Supervised pretraining optimizes per-step classification loss, whereas the quality evaluation is defined on the final graphs. Small errors can compound over steps, causing geometric and topological failures. In addition, IoU is non-differentiable because it depends on decoding and rendering an entire trajectory. We therefore use RL fine-tuning to optimize an outcome-level graph objective under the same rollout dynamics used at test time.

\paragraph{RL with terminal graph reward.} We initialize $\pi_\theta$ from TraversRL-Pre, and keep a copy as frozen reference policy $\pi_{\theta_0}$. For each training sample, we draw a group of $K$ rollouts $\{\tau_i\}_{i=1}^K$ using the same decoding configuration as inference. Each trajectory $\tau_i=(a_{i,1},\ldots,a_{i,T_i})$ induces a predicted edge set $\hat{E}_i$ and receives a terminal, graph-level reward $R_i=R(E,\hat{E}_i)$, which is buffered IoU in our experiments. We compute standardized group-relative advantages:
\begin{equation}
\bar{R}=\frac{1}{K}\sum_{i=1}^K R_i,
\qquad
A_i=\frac{R_i-\bar{R}}{\mathrm{std}(\{R_j\}_{j=1}^K)+\epsilon}.
\end{equation}
We update $\pi_\theta$ to increase the likelihood of actions from higher-advantage rollouts while constraining rapid drift from the frozen reference $\pi_{\theta_0}$ via a KL penalty. Our RL fine-tuning objective is defined as:
\begin{equation}
\begin{aligned}
\mathcal{L}_{\text{global}}(\theta)
&=
-\frac{1}{K}\sum_{i=1}^K \frac{A_i}{T_i}\sum_{t=1}^{T_i}\log \pi_\theta(a_{i,t}\mid s_{i,t})\\
&+\beta\cdot \frac{1}{K}\sum_{i=1}^K \frac{1}{T_i}\sum_{t=1}^{T_i}
D_{\mathrm{KL}}\!\left(\pi_\theta(\cdot\mid s_{i,t}) \,\Vert\, \pi_{\theta_0}(\cdot\mid s_{i,t})\right)
\label{eq:global_rl}
\end{aligned}
\end{equation}
,where $s_{i,t}$ is the rollout state at step $t$ (local image and canvas crops), $\beta$ is the KL coefficient, and $1/T_i$ normalizes trajectories of different lengths. We compute the full-distribution KL at each step from the policy and reference logits.


\subsection{RL Fine-Tuning with Local Stepwise Reward (TraversRL-Local)}\label{sec:grpo_local}

\paragraph{Motivation.} TraversRL-Global aligns fine-tuning with the final objective, but it assigns a single rollout-level advantage $A_i$ to all actions within a trajectory, leading to coarse credit assignment. In practice, however, a rollout often contains a mixture of locally correct and incorrect steps, yet all steps are reinforced (or penalized) uniformly. As a result, terminal-only reward can improve global overlap, but under-optimizes fine-grained geometric choices (e.g., selecting a nearby but incorrect direction bin that yields similar buffered overlap).

\begin{figure}[!t]
    \centering
    \includegraphics[width=\textwidth]{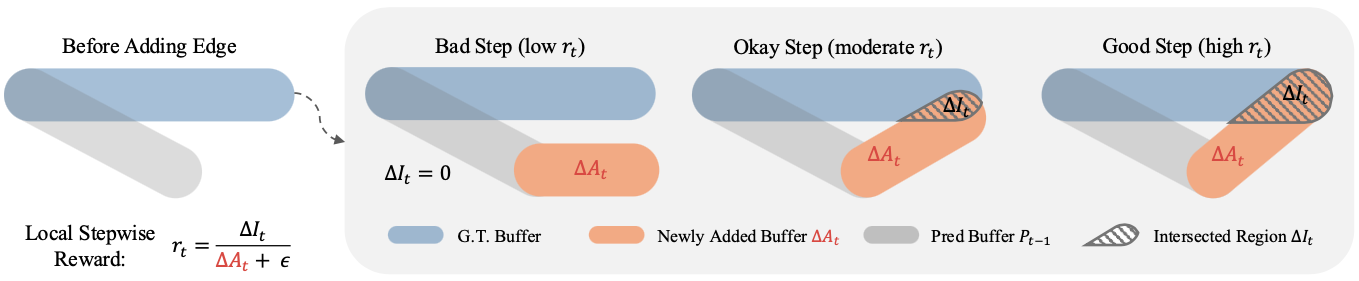}
    \caption{\textbf{Local stepwise reward}. We compare the buffered ground-truth region $G$ (blue) with the current predicted buffer $P_{t-1}$ (gray). Adding a new edge at step $t$ contributes newly added buffered area $\Delta A_t$ (orange), of which the overlapped portion with ground-truth is $\Delta I_t$. We define the stepwise reward $r_t=\Delta I_t/(\Delta A_t+\epsilon)$: it is high when the added edge aligns better with ground-truth (good step), moderate under partial alignment, and near zero when the edge adds off-target geometry ($\Delta I_t=0$).} \label{fig:local_reward}
\end{figure}

\paragraph{Local stepwise reward}. To inject a localized training signal without discarding the stable terminal objective, we define a stepwise reward that quantifies the local geometric quality of each newly added edge and use it for advantage shaping. Let $P_t$ be the buffered union of predicted edges up to step $t$ and let $G$ be the buffered union of ground-truth edges. Define predicted area $A_t=\mathrm{area}(P_t)$ and overlap area $I_t=\mathrm{area}(P_t\cap G)$. With a new edge, we compute $\Delta A_t = A_t - A_{t-1}$ and $\Delta I_t = I_t - I_{t-1}$, formulating the stepwise reward as:
\begin{equation}
r_t \;=\; \frac{\Delta I_t}{\Delta A_t + \epsilon},
\label{eq:eff_reward}
\end{equation}
with $r_t=0$ for non-edge actions (e.g., \textsc{Stop}). Intuitively, $r_t$ measures what fraction of newly generated buffered area contributes to ground-truth overlap; edges that add mostly off-target geometry receive near-zero reward (Fig~\ref{fig:local_reward}).

\begin{figure}[!t]
    \centering
    \includegraphics[width=\columnwidth]{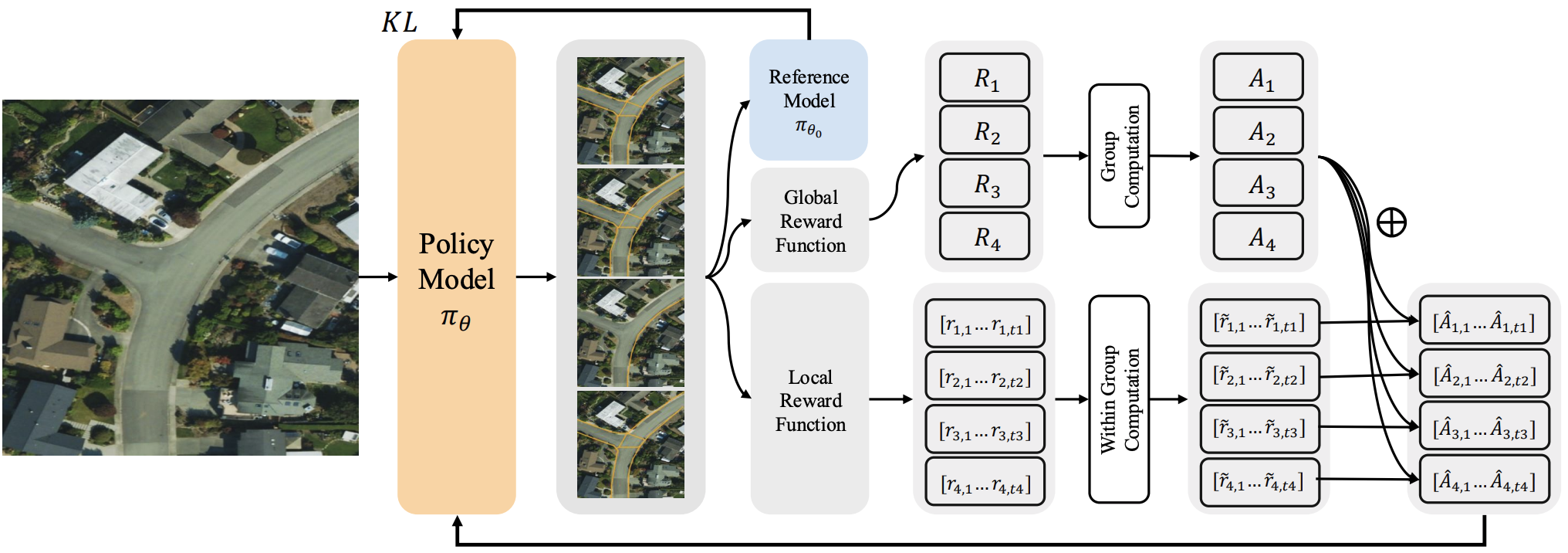}
    \caption{\textbf{RL fine-tuning with local stepwise reward shaping (TraversRL-Local).} For each intersection, we sample $K$ rollouts (here $K{=}4$) from $\pi_\theta$, producing action sequences and generated graphs. A global reward assigns each rollout a terminal reward $R_i$, which is converted into group-relative advantages $\{A_i\}_{i=1}^K$. In parallel, a local reward computes per-step rewards $\{r_{i,t}\}$, which are normalized within each rollout to obtain $\{\tilde{r}_{i,t}\}$ and combined with $A_i$ to form shaped advantages $\hat{A}_{i,t}=A_i+\lambda \tilde{r}_{i,t}$. The policy is updated using $\hat{A}_{i,t}$ together with KL$(\pi_\theta\Vert\pi_{\theta_0})$; only $\pi_\theta$ is updated, while $\pi_{\theta_0}$ is frozen. Without local shaping (TraversRL-Global), the update reduces to $\hat{A}_{i,t}=A_i$.}
    \label{fig:traversrl}
\end{figure}

\paragraph{Additive advantage shaping.}
As in Sec.~\ref{sec:grpo_global}, we sample $K$ rollouts and compute the terminal advantages $A_i$ from the final rewards $R_i$. We then calculate and normalize the stepwise rewards within each individual rollout:
\begin{equation}
\tilde{r}_{i,t}
\;=\;
\frac{r_{i,t}-\mu_i}{\sigma_i+\epsilon},
\quad
\mu_i=\frac{1}{T_i}\sum_{t=1}^{T_i} r_{i,t},
\quad
\sigma_i=\mathrm{std}\!\left(\{r_{i,t}\}_{t=1}^{T_i}\right),
\label{eq:local_norm}
\end{equation}
We 
define an additively shaped stepwise advantage:
\begin{equation}
\hat{A}_{i,t} \;=\; A_i \;+\; \lambda\,\tilde{r}_{i,t},
\end{equation}
where $\lambda$ controls the strength of shaping. The additive formulation preserves the outcome-level signal, while simultaneously encouraging locally high-quality steps even within imperfect rollouts. Since both $A_i$ and $\tilde{r}_{i,t}$ are standardized, their magnitudes are comparable, making the shaping term well-scaled in practice. We then substitute the uniform rollout-level weighting in Eq.~\eqref{eq:global_rl} with $\hat{A}_{i,t}$ while keeping the same KL regularizer to the frozen reference:
\begin{equation}
\begin{aligned}
\mathcal{L}_{\text{local}}(\theta)
&=
-\frac{1}{K}\sum_{i=1}^K \frac{1}{T_i}\sum_{t=1}^{T_i}
\hat{A}_{i,t}\log \pi_\theta(a_{i,t}\mid s_{i,t})\\
&+\beta\cdot \frac{1}{K}\sum_{i=1}^K \frac{1}{T_i}\sum_{t=1}^{T_i}
D_{\mathrm{KL}}\!\left(\pi_\theta(\cdot\mid s_{i,t}) \,\Vert\, \pi_{\theta_0}(\cdot\mid s_{i,t})\right)
\label{eq:local_rl}
\end{aligned}
\end{equation}
This procedure is visually demonstrated in Fig.~\ref{fig:traversrl}.

\paragraph{Note:} A natural alternative is to replace the terminal advantage with stepwise advantages normalized across rollouts at each step. However, we found this to be empirically unstable in our task. Step rewards are highly sensitive to small geometric perturbations and vary substantially across rollouts and step indices due to different rollout lengths and exploration paths. This high variance makes cross-rollout, per-step normalization noisy. In practice, using purely stepwise advantages causes rapid drift from the reference policy (KL increases sharply early in training) and leads to collapse, whereas advantage shaping preserves the stability of terminal-reward RL while still providing informative local credit.

%% file: sections/experiment.tex
We evaluate TraversRL in both in-domain and zero-shot transfer settings. Sec.~\ref{sec:datasets} details the evaluation datasets, Sec.~\ref{sec:backbones} outlines the visual backbones and baseline models, and Sec.~\ref{sec:training} specifies the training and evaluation protocols.

\subsection{Datasets}\label{sec:datasets}

We evaluate across three urban-intersection datasets: \textbf{\circlednum{1} WashingtonInter (in-domain)}. A curated dataset of three- and four-way urban intersections from 13 cities in Washington State. We first extract intersection bounding boxes from OpenStreetMap and prune overlapping regions to avoid spatial train--test leakage. We then retrieve the corresponding pedestrian pathway annotations from our internal database and discard scenes whose ground-truth graph is disconnected. After a final manual visual inspection to remove severe aerial-to-map misalignments, the dataset contains 2,287 training scenes, 144 validation scenes, and 426 test scenes. \textbf{\circlednum{2} PathwayBench, Seattle (zero-shot)}~\cite{zhang2024pathwaybench}. We evaluate 342 intersection scenes from Seattle. Since PathwayBench does not provide training and validation splits, we evaluate in a \emph{zero-shot} setting: models are trained on WashingtonInter and applied directly to PathwayBench without fine-tuning. We ensure that there is no overlap intersection with the WashingtonInter data. \textbf{\circlednum{3} PathwayBench, Washington, D.C. (zero-shot)}~\cite{zhang2024pathwaybench}. 589 intersection scenes from Washington, D.C., evaluated under the same zero-shot setting.

PathwayBench uses different scene extents and spatial resolutions (an effective zoom level of roughly 19, compared to approximately 20 in WashingtonInter). Therefore, we refer to all PathwayBench evaluations as zero-shot transfer.

\subsection{Backbones and Baselines}\label{sec:backbones}

We initiate TraversRL with three ImageNet-pretrained visual backbones: ResNet-18~\cite{he2016deep}, ViT-S~\cite{dosovitskiy2020image}, and Swin-T~\cite{liu2021swin}. All are adapted to 4-channel inputs by modifying the first projection layer to accept an additional canvas channel while retaining pretrained RGB weights (the extra channel is initialized to zero). ViT uses $224\times224$ inputs, while ResNet and Swin use $256\times256$. We compare against Tile2Net~\cite{hosseini2023mapping}, a state-of-the-art pipeline that predicts sidewalks, crosswalks, and footpaths from aerial imagery and converts them into pedestrian networks. 

We also evaluate two road-oriented graph extraction methods, VecRoad~\cite{tan2020vecroad} and NETracer~\cite{liu2025netracer}, to test whether existing road-tracing pipelines transfer to pedestrian pathways. For NETracer, we convert pathway annotations into the required raster/SWC and seed/centerline inputs. For VecRoad, we convert pathway graphs into OSM-style supervision, derive junction labels from graph topology, and tune key tracing parameters. These baselines provide graph-based comparisons while highlighting the domain gap: pedestrian pathways are typically narrower, shorter, more occluded, and have subtler junction cues than roads.

\subsection{Training \& Evaluation}\label{sec:training}

\textbf{Supervised pretraining.} For each backbone, we train a supervised reference model following Sec.~\ref{sec:pretraining}. Unless otherwise noted, we apply localization error with $\eta=0.02$ (in normalized coordinates), and optimize with AdamW (learning rate $1\times10^{-4}$, weight decay $1\times10^{-4}$, gradient clipping at 1.0). Models are trained for 50 epochs with gradient accumulation over 16 node-expansions, with max\_steps=200. For ResNet-18 backbones, BatchNorm statistics are frozen. We select the best checkpoint by validation loss and use it to initialize RL fine-tuning.

\textbf{RL fine-tuning.} We fine-tune each pretrained checkpoint using GRPO (Secs.~\ref{sec:grpo_global}, \ref{sec:grpo_local}). Unless otherwise noted, we use AdamW (learning rate $3\times10^{-5}$, weight decay $1\times10^{-4}$, gradient clipping at 1.0), train for 15 epochs with gradient accumulation over 8 graphs, and set the KL coefficient to $\beta=0.05$. For each training graph, we sample a group of $K=4$ rollouts, capping the maximum rollout length at 200 steps. 
Rewards use a buffer radius $r=0.01$ ($\approx$ 1 meter). For the local shaping variant, we set $\lambda=0.1$ and clip $\tilde{r}_{i,t}$ to be within $[-3,3]$ for stability. We select the best RL checkpoint by validation buffered IoU.

We evaluate each method over five randomly sampled start nodes and report averages. We use two metrics: \textbf{\circlednum{1} Buffered geometry IoU}, computed between the predicted and ground-truth edge sets by buffering each polyline with a 1-meter radius and calculating the IoU of the resulting regions. 
and \textbf{\circlednum{2} Traversability similarity (TravSim)}~\cite{zhang2024pathwaybench}, a routability-oriented metric that compares boundary-to-boundary connectivity within local tiles. A boundary pair is considered traversable if a valid path exists within the clipped tile connecting a terminating node on one boundary to a terminating node on another. Each tile thus induces a set of traversable boundary pairs for both prediction and ground-truth, and TravSim is the Jaccard similarity between these two sets, averaged over tiles. This metric does not require node/edge correspondence and directly measures whether the generated graph affords the same traversal possibilities as the ground-truth within intersection-scale regions.

%% file: sections/results.tex
In this section, we present detailed generation performances. We structure these findings into \textit{takeaways} to facilitate understanding,  denoted as \textbf{T\#}.

\subsection{(T1) Supervised Policies Strongly Outperform the Baseline}\label{sec:t1}

Table~\ref{tab:quantitative_results} compares pretrained checkpoints (TraversRL-Pre) against Tile2Net across three datasets. Even before RL fine-tuning, the pretrained iterative graph generator yields substantially higher geometric overlap and connectivity.

\textbf{\circlednum{1} WashingtonInter (in-domain).} Tile2Net achieves 0.264 IoU and 0.114 TravSim on average. In contrast, TraversRL-Pre improves both metrics for every backbone, reaching 0.531 IoU and 0.605 TravSim with Swin-T. Averaged across backbones, TraversRL-Pre exceeds Tile2Net by 0.258 IoU and 0.481 TravSim. The large TravSim gain suggests the iterative formulation better preserves intersection topology than raster-to-network post-processing, where small geometric artifacts can disconnect crossings and sidewalk continuations. 

\textbf{\circlednum{2} PathwayBench (zero-shot transfer).} The same trend holds under distribution shift. On Seattle data, TraversRL-Pre outperforms Tile2Net by an average of 0.026 IoU and 0.215 TravSim. On D.C. data, TraversRL-Pre improves TravSim by 0.196 on average, while achieving competitive IoU: Swin-T exceeds Tile2Net in IoU (0.231 vs. 0.226), whereas ResNet-18 and ViT-S fall marginally behind. Overall, cross-city transfer is more robust for connectivity. 

\textbf{\circlednum{3} Road-oriented methods.} Road-oriented graph methods do not transfer well to pedestrian pathway generation: NETracer produced fragmented graphs likely because its road-tuned tracing rules prematurely reject valid but subtle pathway continuations. VecRoad produced near-empty outputs because it relies on high-confidence junction maps, while our generated pathway junctions are topologically valid but sparse and visually weak. These results suggest that released road-tracing pipelines cannot be directly applied to pedestrian pathways without substantial method-specific redesign.

\input{tables/quantitative_results}


\subsection{(T2) RL Fine-Tuning Consistently Improves Performance Across Backbones and Datasets}\label{sec:t2}

Table~\ref{tab:quantitative_results} shows that RL fine-tuning with an outcome-level reward improves performance across backbones and datasets, despite the already strong TraversRL-Pre baseline. With the strict 1 m buffer, IoU becomes highly sensitive to geometric misalignment, making improvements more indicative of true precision.

\textbf{\circlednum{1} Global reward improves graph quality.} With a terminal graph-level reward, TraversRL-Global improves IoU for every backbone on every dataset over TraversRL-Pre. Averaged across backbones, the relative IoU gains are 5.1\% on WashingtonInter, 5.6\% on Seattle, and 5.2\% on D.C. TravSim also increases by 0.6\%, 3.0\%, and 0.5\%, respectively. These consistent trends indicate that optimizing a terminal objective aligns the policy toward decisions that improve both geometric overlap and connectivity patterns.

\begin{figure}[!t]
    \centering
    \includegraphics[width=\columnwidth]{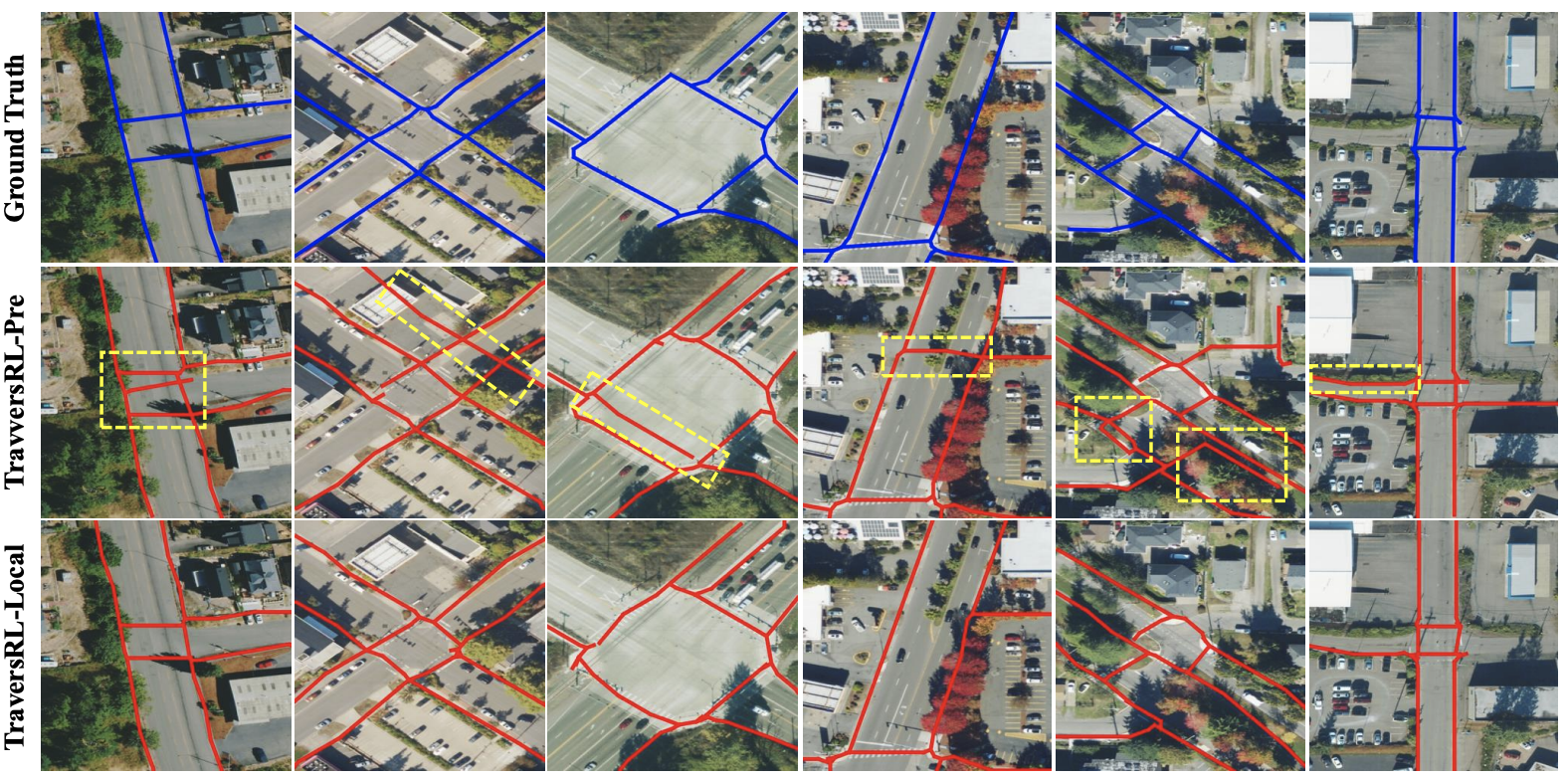}
    \caption{\textbf{RL fine-tuning produces cleaner, sparser graphs.}
    Qualitative examples comparing the ground-truth graph (top, blue) to predictions from TraversRL-Pre (middle, red, ``messy'') and TraversRL-Local (bottom, red, ``cleaner''). Yellow dashed boxes highlight artifacts of the supervised policy --- redundant branches, spurious short connectors, and noisy junction geometry ---that are pruned after RL fine-tuning.}
    \label{fig:clean_graphs}
\end{figure}

\textbf{\circlednum{2} Local reward shaping adds further gains.} Relative to TraversRL-Global, TraversRL-Local yields additional IoU gains of 1.6\% (WashingtonInter), 4.1\% (Seattle), and 2.1\% (D.C.), with one minor exception (ViT-S on D.C.). Relative to TraversRL-Pre, these gains surge to 6.7\%, 9.8\%, and 7.5\%. Local shaping also consistently improves TravSim, by 1.5\% on WashingtonInter, 2.4\% on Seattle, and 1.0\% on D.C. These results support the intended mechanism of the local reward: finer credit assignment encourages locally aligned edge placement while preserving stable terminal-reward optimization.

Additionally, the standard deviations in Table~\ref{tab:quantitative_results} are generally small across five random seeds. In most settings, the RL gains in IoU are comparable to, or larger than, the seed-to-seed variation, indicating that the gains are not an artifact of a favorable seed but reflect a systematic improvement in graph quality.

\input{tables/avgdeg_travsim}

\textbf{\circlednum{3} TraversRL-Local recovers subtle missing paths.} Recall that TravSim compares the set of traversable boundary-to-boundary pairs induced by the predicted vs. ground-truth graphs within each tile  (Sec.~\ref{sec:training}). Let $S_{\text{pred}}$ and $S_{\text{gt}}$ denote these per-tile sets of traversable boundary-pair relations for the prediction and ground-truth, respectively. We decompose their disagreement into: $\mathrm{TP}=\lvert S_{\text{pred}}\cap S_{\text{gt}}\rvert$,
$\mathrm{FP}=\lvert S_{\text{pred}}\rvert-\mathrm{TP}$, and $\mathrm{FN}=\lvert S_{\text{gt}}\rvert-\mathrm{TP}$. On WashingtonInter, averaging per-tile set sizes over tiles and then over seeds/backbones, TraversRL-Pre predicts approximately 1.86 boundary-pair relations per tile, with $\mathrm{TP}\approx 1.55$, $\mathrm{FP}\approx 0.31$, and $\mathrm{FN}\approx 1.09$ (i.e., missed connectivity relations dominate spurious ones by $\sim$3.5$\times$). RL fine-tuning primarily reduces this missed-connectivity term: TraversRL-Local increases $\mathrm{TP}$ from $1.55\rightarrow 1.60$ ($+3.5\%$) and reduces $\mathrm{FN}$ from $1.09\rightarrow 1.04$ ($-5.0\%$). Consistent with this, at the scene level TravSim improves in $59\%$ of intersections; among those improved scenes, $92\%$ exhibit increased $\mathrm{TP}$ and $91\%$ exhibit reduced $\mathrm{FN}$. Overall, RL improves TravSim by recovering missing ground-truth paths.


\subsection{(T3) TraversRL Produces Cleaner, Sparser Graphs While Preserving Connectivity.}\label{sec:t3}

Beyond buffered IoU, we analyze how RL changes graph structure. Following PathwayBench, we compute all graph statistics within each tile on the clipped predicted graph and then average over tiles. We report: (1) average node degree (AvgDeg) as a proxy for graph density/cleanliness (lower typically indicates fewer branches) and (2) TravSim as a proxy for whether the predicted graph preserves ground-truth connectivity.  AvgDeg alone cannot distinguish clean pruning from missing edges, so we interpret it jointly with TravSim.

\paragraph{RL reduces graph density while improving connectivity agreement.} A consistent pattern emerges across all datasets (Table~\ref{tab:avgdeg_travsim}): RL fine-tuning yields sparser graphs while maintaining or improving connectivity. Averaged across backbones, TraversRL-Global reduces AvgDeg relative to TraversRL-Pre by 3.0\% on WashingtonInter, 5.7\% on Seattle, and 4.5\% on D.C., while increasing TravSim by 0.6\%, 3.1\%, and 0.6\%, respectively. TraversRL-Local strengthens this trend, further reducing AvgDeg (5.1\%, 9.6\%, 6.0\%) while improving TravSim (2.0\%, 5.7\%, 1.6\%). Fig.~\ref{fig:clean_graphs} shows the same effect qualitatively: RL suppresses redundant edges and noisy branches while preserving intersection structure. 


\subsection{(T4) Buffer-Size Ablation: Stricter Buffers Reveal Larger Gains.}\label{sec:t4}

To study metric sensitivity, we re-evaluated predictions with wider buffers. On WashingtonInter dataset, averaged across backbones and seeds, TraversRL-Global improves IoU over TraversRL-Pre by 2.1\% at 4m, which grows to 3.1\% at 2m and 5.2\% at 1m. The same pattern holds for TraversRL-Local, whose improvements increase by 2.5\%, 4.0\% and 6.8\% respectively. These observations show that outcome-level RL fine-tuning improves geometric precision in a way that becomes increasingly visible as the overlap criterion tightens, with local shaping providing the strongest improvements under strict alignment. The effect is strongest for local shaping: relative to global-reward RL, TraversRL-Local gains widen as the buffer shrinks, consistent with the reward encouraging locally aligned edge extensions and discouraging off-target geometry.


\subsection{(T5) Action-Space Ablation: Finer Actions Improve In-Domain Performance but Show Mixed Transfer.}\label{sec:t5}

We study sensitivity to action-space granularity by fixing the number of distance bins to $10$ and varying the number of angle bins to $18$ and $72$. We run this ablation with ResNet-18 and TraversRL-Local. We observe that: (1) Reducing the action space to $|\mathcal{A}|=181$ consistently hurts performance, lowering IoU/TravSim from $0.540/0.605$ to $0.448/0.581$ on WashingtonInter, from $0.199/0.403$ to $0.174/0.385$ on Seattle, and from $0.227/0.513$ to $0.195/0.503$ on D.C. (2) Increasing the action space to $|\mathcal{A}|=721$ improves in-domain performance to $0.569/0.618$, but shows mixed transfer: Seattle slightly drops to $0.198/0.383$, while D.C. improves to $0.237/0.520$. These results suggest that finer actions can reduce angular discretization error, but may also overfit on the source data and increase sensitivity to dataset-specific scale or noise. Overall, the default $|\mathcal{A}|=361$ provides a reasonable tradeoff between precision and cross-domain robustness.

%% file: tables/quantitative_results.tex
\begin{table}[t]
    \centering
    \scriptsize
    \setlength{\tabcolsep}{2.25pt}
    \renewcommand{\arraystretch}{1.25}
    \begin{tabular}{c c c |cc |cc |cc}
        \toprule
        \multicolumn{2}{c}{\textbf{Model}} & \textbf{Stage}& \multicolumn{2}{c|}{\textbf{WashingtonInter}} & \multicolumn{2}{c|}{\textbf{Seattle (PB)}} & \multicolumn{2}{c}{\textbf{D.C. (PB)}}\\
        \cmidrule(lr){4-5}\cmidrule(lr){6-7} \cmidrule(lr){8-9}
        & & & \textbf{IoU} & \textbf{TravSim} & \textbf{IoU} & \textbf{TravSim} & \textbf{IoU} & \textbf{TravSim} \\
        \midrule
        \multicolumn{2}{c}{VecRoad} & -- & $<0.001$ & $<0.001$ & $<0.001$ & $<0.001$ & $<0.001$ & $<0.001$ \\
        \multicolumn{2}{c}{NETracer} & -- & 0.098 & 0.005 & 0.080 & 0.003 & 0.007 & 0.002 \\
        \multicolumn{2}{c}{Tile2Net} & -- & 0.264$_{\pm0.046}$ & 0.114$_{\pm0.017}$ & 0.169$_{\pm0.021}$ & 0.175$_{\pm0.037}$ & 0.226$_{\pm0.018}$ & 0.317$_{\pm0.023}$ \\
        \midrule 
        \multirow{9}{*}{\textbf{\rotatebox{90}{TraversRL}}}
        & \multirow{3}{*}{\rotatebox{90}{\tiny{ResNet18}}}
        & pre & 0.510$_{\pm0.003}$ & 0.586$_{\pm0.005}$ & 0.186$_{\pm0.003}$ & 0.386$_{\pm0.003}$ & 0.212$_{\pm0.003}$ & 0.506$_{\pm0.004}$ \\
        & & global & 0.539$_{\pm0.001}^{\dagger}$ & 0.591$_{\pm0.008}^{\dagger}$ & 0.189$_{\pm0.002}^{\dagger}$ & 0.387$_{\pm0.004}^{\dagger}$ & 0.223$_{\pm0.002}$ & 0.507$_{\pm0.003}^{\dagger}$ \\
        & & local & 0.540$_{\pm0.002}^{*}$ & 0.605$_{\pm0.004}^{*}$ & 0.199$_{\pm0.003}^{*}$ & 0.403$_{\pm0.002}^{*}$ & 0.227$_{\pm0.002}^{*}$ & 0.513$_{\pm0.002}^{*}$ \\
        \cmidrule(lr){2-9}
        & \multirow{3}{*}{{\rotatebox{90}{\tiny{ViT-S}}}}
        & pre & 0.525$_{\pm0.003}$ & 0.593$_{\pm0.005}$ & 0.194$_{\pm0.003}$ & 0.382$_{\pm0.006}$ & 0.208$_{\pm0.002}$ & 0.499$_{\pm0.002}$ \\
        & & global & 0.548$_{\pm0.003}^{\dagger}$ & 0.597$_{\pm0.003}^{\dagger}$ & 0.211$_{\pm0.002}^{\dagger}$ & 0.401$_{\pm0.005}^{\dagger}$ & 0.226$_{\pm0.001}^{\dagger}$ & 0.506$_{\pm0.004}^{\dagger}$ \\
        & & local & 0.548$_{\pm0.001}^{\dagger}$ & 0.605$_{\pm0.005}^{*}$ & 0.214$_{\pm0.003}^{*}$ & 0.411$_{\pm0.003}^{*}$ & 0.221$_{\pm0.002}$ & 0.515$_{\pm0.004}^{*}$ \\
        \cmidrule(lr){2-9}
        & \multirow{3}{*}{\rotatebox{90}{{\tiny{Swin-T}}}}
        & pre & 0.531$_{\pm0.004}$ & 0.605$_{\pm0.006}$ & 0.205$_{\pm0.004}$ & 0.401$_{\pm0.004}$ & 0.231$_{\pm0.002}$ & \textbf{0.535}$_{\pm0.004}$ \\
        & & global & 0.559$_{\pm0.002}^{\dagger}$ & 0.607$_{\pm0.004}^{\dagger}$ & 0.218$_{\pm0.001}^{\dagger}$ & 0.416$_{\pm0.004}^{\dagger}$ & 0.236$_{\pm0.001}^{\dagger}$ & \textbf{0.535}$_{\pm0.004}$ \\
        & & local & \textbf{0.584}$_{\pm0.003}^{*}$ & \textbf{0.611}$_{\pm0.008}^{*}$ & \textbf{0.230}$_{\pm0.005}^{*}$ & \textbf{0.418}$_{\pm0.007}^{*}$ & \textbf{0.252}$_{\pm0.003}^{*}$ & \textbf{0.535}$_{\pm0.003}$ \\
        \bottomrule
    \end{tabular}
    \vspace{0.25cm}
    \caption{\textbf{Quantitative results} on WashingtonInter (in-domain) and PathwayBench test cities (zero-shot transfer), reported as mean $\pm$ std over five seeds. \textbf{pre}, \textbf{global}, and \textbf{local} denote supervised pretraining, RL fine-tuning with a terminal graph-level reward, and RL fine-tuning with additional local stepwise reward shaping, respectively. \textbf{Bold} indicates the best value within each dataset/metric column. $^{\dagger}$ denotes an improvement over both baselines and the corresponding \textbf{pre} checkpoint (same backbone). $^{*}$ denotes an improvement over the corresponding \textbf{global} checkpoint (same backbone).}
    \label{tab:quantitative_results}
\end{table}

%% file: tables/avgdeg_travsim.tex
\begin{table}[t] 
    \centering
    \scriptsize
    \setlength{\tabcolsep}{4pt} 
    \renewcommand{\arraystretch}{1.25}
    \begin{tabular}{c|ccc|ccc} 
        \toprule & \multicolumn{3}{c|}{AvgDeg$\downarrow$} & \multicolumn{3}{c}{TravSim$\uparrow$}\\ \textbf{Dataset} & \textbf{Pre} & \textbf{Global ($\Delta$)} & \textbf{Local ($\Delta$)} & \textbf{Pre} & \textbf{Global ($\Delta$)} & \textbf{Local ($\Delta$)} \\ 
        \midrule WashingtonInter & 0.429 & 0.416 (-3.0\%) & 0.407 (-5.1\%) & 0.595 & 0.599 (+0.6\%) & 0.607 (+2.0\%) \\ Seattle (PB) & 0.574 & 0.541 (-5.7\%) & 0.519 (-9.6\%) & 0.389 & 0.401 (+3.1\%) & 0.411 (+5.7\%) \\ D.C.\ (PB) & 0.553 & 0.528 (-4.5\%) & 0.520 (-6.0\%) & 0.513 & 0.516 (+0.6\%) & 0.521 (+1.6\%) \\ \bottomrule 
    \end{tabular} 
    \vspace{0.25cm}
    \caption{\textbf{Graph cleanliness and connectivity agreement} averaged over backbones and five seeds. AvgDeg denotes average node degree (lower indicates sparser graphs), and TravSim denotes traversability similarity (higher indicates closer agreement with ground-truth boundary connectivity). For Global and Local, we report the relative change compared to Pre on the same dataset.}
    \label{tab:avgdeg_travsim} 
\end{table} 

%% file: sections/conclusion.tex
We introduced TraversRL, a vision-conditioned iterative generator for constructing pedestrian pathway graphs from aerial imagery.  TraversRL represents state with a local image crop and a generation canvas encoding the partial graph, and predicts discretized direction–distance actions tailored to short pedestrian segments. Across backbones and datasets, RL fine-tuning consistently improves buffered IoU and connectivity agreement, while also simplifying graph structure, demonstrating effectiveness for long-horizon pedestrian network generation.

Our work also suggests several directions for improvement. (1) Scale variation. TraversRL uses a fixed crop size in normalized coordinates, so the effective physical scale varies across datasets and scene extents. This can obscure thin pathways and exacerbate domain shift. Scale-aware inputs could improve cross-city robustness. (2) Start-node initialization. For evaluation, rollouts are initialized from a start node on the ground-truth graph. In deployment, this information is unavailable, so a practical system will require a seeding mechanism, e.g., predicting likely start node from imagery from a segmentation prior. (3) Coverage of disconnected components. Some scenes contain multiple disconnected pathway components. A single-seed rollout can only recover the component reachable from that seed. Full coverage requires multi-seed inference and a strategy for proposing and filtering seeds in unexplored regions.

Ultimately, we envision an extended TraversRL model to ``walk the city'' and produce high-quality de novo pedestrian networks without human intervention.